\documentclass[10pt,twocolumn,letterpaper]{article}

\usepackage{iccv}
\usepackage{times}
\usepackage{epsfig}
\usepackage{graphicx}
\usepackage{amsmath}
\usepackage{amssymb}

\usepackage{mathptmx}
\usepackage{mathrsfs}
\usepackage{xcolor}
\usepackage{makecell}
\usepackage{mathtools}
\usepackage{bbm}

\usepackage{booktabs}
\usepackage{microtype}
\usepackage{xspace}
\usepackage{multirow}
\usepackage{soul}
\usepackage{tabularx}
\usepackage{enumitem}

\usepackage[pagebackref=true,breaklinks=true,letterpaper=true,colorlinks,bookmarks=false]{hyperref}

\iccvfinalcopy 

\def\iccvPaperID{1027} 

\newcommand{\mypara}[1]{\vspace{1mm}\noindent \textbf{#1}}
\newcommand{\titlecap}[2]{\textbf{#1} #2}

\providecommand{\ie}{\textit{i.e.}\xspace}
\providecommand{\eg}{\textit{e.g.}\xspace}
\providecommand{\etc}{\textit{etc.}\xspace}

\begin{document}

\title{Where2Act: From Pixels to Actions for Articulated 3D Objects}


\author{
Kaichun Mo$^{*1}$ \, Leonidas Guibas$^{1}$ \, Mustafa Mukadam$^{2}$  \, Abhinav Gupta$^{2}$ \, Shubham Tulsiani$^{2}$
\vspace{0.2cm} \\ 
$^{1}$Stanford University \, $^{2}$Facebook AI Research
\vspace{0.2cm} \\
\url{https://cs.stanford.edu/~kaichun/where2act}
}

\maketitle

\renewcommand*{\thefootnote}{\fnsymbol{footnote}}
\footnotetext[1]{The majority of the work was done while Kaichun Mo was a research intern at Facebook AI Research.}
\renewcommand*{\thefootnote}{\arabic{footnote}}


\label{abs}
\begin{abstract}
One of the fundamental goals of visual perception is to allow agents to meaningfully interact with their environment. In this paper, we take a step towards that long-term goal -- we extract highly localized actionable information related to elementary actions such as pushing or pulling for articulated objects with movable parts. For example, given a drawer, our network predicts that applying a pulling force on the handle opens the drawer. We propose, discuss, and evaluate novel network architectures that given image and depth data, predict the set of actions possible at each pixel, and the regions over articulated parts that are likely to move under the force. We propose a learning-from-interaction framework with an online data sampling strategy that allows us to train the network in simulation (SAPIEN) and generalizes across categories. 
Check the website for code and data release.

\end{abstract}

\section{Introduction}
\label{sec:intro}


\begin{figure}[t!]
    \centering
    \includegraphics[width=\linewidth]{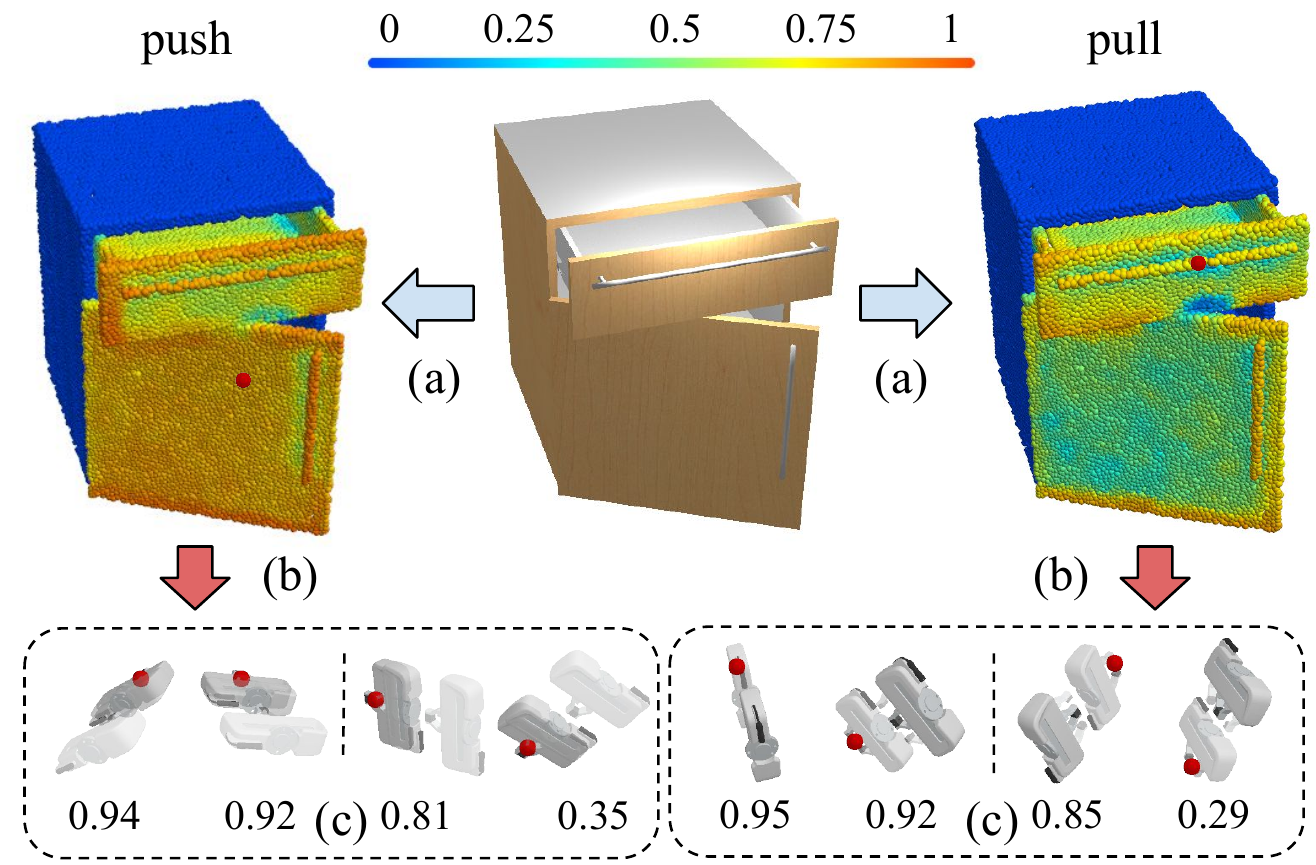}
    \caption{\titlecap{The Proposed Where2Act Task.}{Given as input an articulated 3D object, we learn to propose the actionable information for different robotic manipulation primitives (\eg \textit{pushing}, \textit{pulling}): (a) the predicted actionability scores over pixels; (b) the proposed interaction trajectories, along with (c) their success likelihoods, for a selected pixel highlighted in red. We show two high-rated proposals (left) and two with lower scores (right) due to interaction orientations and potential robot-object collisions. }}
    \label{fig:teaser}
    \vspace{-4mm}
\end{figure}

We humans interact with a plethora of objects around us in our daily lives. What makes this possible is our effortless understanding of \emph{what} can be done with each object, \emph{where} this interaction may occur, and precisely \emph{how} our we must move to accomplish it -- we can pull on a handle to open a drawer, push anywhere on a door to close it, flip a switch to turn a light on, or push a button to start the microwave. Not only do we understand what actions will be successful, we also intuitively know which ones will not \eg pulling out a remote's button is probably not a good idea! In this work, our goal is to build a perception system which also has a similar understanding of general objects \ie given a novel object, we want a system that can infer the myriad possible interactions\footnote{Gibson proposed the idea of affordances -- opportunities of interaction. Classical notion of object affordance involves consideration of agent's morphology. Our interactions are more low-level actions.} that one can perform with it. 


The task of predicting possible interactions with objects is one of central importance in both, the robotics and the computer vision communities. In robotics, the ability to predict feasible and desirable actions  (\eg a drawer can be pulled out) can help in motion planning, efficient exploration and interactive learning (sampling successful trials faster). On the other hand, the computer vision community has largely focused on inferring semantic labels (\eg part segmentation, keypoint estimation) from visual input, but such passively learned representations provide limited understanding. More specifically, passive learning falls short on the ability of agents to perform actions, learn prediction models (forward dynamics) or even semantics in many cases (categories are more than often defined on affordances themselves!). Our paper takes a step forward in building a common perception system across diverse objects, while creating its own supervision about what actions maybe successful by actively interacting with the objects.


The first question we must tackle is how one can parametrize the predicted action space. We note that any long-term interaction with an object can be considered as a sequence of short-term `atomic' interactions like pushing and pulling. We therefore limit our work to considering the plausible short-term interactions that an agent can perform given the current state of the object. Each such atomic interaction can further be decomposed into \textit{where} and \textit{how} \eg where on the cabinet should the robot pull (\eg drawer handle or drawer surface) and how should the motion be executed (\eg pull parallel or perpendicular to handle). This observation allows us to formulate our task as one of dense visual prediction. Given a depth or color image of an object, we learn to infer for each pixel/point, whether a certain primitive action can be performed at that location, and if so, how it should be executed.

Concretely, as we illustrate in Figure~\ref{fig:teaser} (a), we learn a prediction network that given an atomic action type, can predict for each pixel: a) an `actionability' score, b) action proposals, and c) success likelihoods. Our approach allows an agent to learn these by simply interacting with various objects, and recording the outcomes of its actions  -- labeling ones that cause a desirable state change as successful. While randomly interacting can eventually allow an agent to learn, we observe that it is not a very efficient exploration strategy. We therefore propose an on-policy data sampling strategy to alleviate this issue -- by biasing the sampling towards actions the agents thinks are likely to succeed.

We use the SAPIEN~\cite{xiang2020sapien} simulator for learning and testing our approach for six types of primitive interaction, covering 972 shapes over 15 commonly seen indoor object categories. We empirically show that our method successfully learns to predict possible actions for novel objects, and does so even for previously unseen categories. 
In summary, our contributions are:
\begin{itemize}
    \vspace{-2mm}
    \item we formulate the task of inferring affordances for manipulating 3D articulated objects by predicting per-pixel action likelihoods and proposals;
    \vspace{-2mm}
    \item we propose an approach that can learn from interactions while using adaptive sampling to obtain more informative samples;
    \vspace{-2mm}
    \item we create benchmarking environments in SAPIEN, and show that our network learns actionable visual representations that generalize to novel shapes and even unseen object categories.
\end{itemize}

\section{Related Works}
\label{sec:related}

\mypara{Predicting Semantic Representations.}
To successfully interact with a 3D object, an agent must be able to `understand' it given some perceptual input. Several previous works in the computer vision community have pursued such an understanding in the form of myriad semantic labels. 
For example, predicting category labels~\cite{wu20153d,chang2015shapenet}, or more fine-grained output such as semantic keypoints~\cite{dutagaci2012evaluation,You_2020_CVPR} or part segmentations~\cite{yi2016scalable,mo2019partnet} can arguably yield more actionable representations \eg allowing one to infer where `handles', or `buttons' \etc are. However, merely obtaining such semantic labels is clearly not sufficient on its own -- an agent must also understand \emph{what} needs to be done (\eg an handle can be `pulled' to open a door), and \emph{how} that action should be accomplished \ie what precise movements are required to `pull open' the specific object considered.

\mypara{Inferring Geometric and Physical Properties.}
Towards obtaining information more directly useful for \emph{how} to act, some methods aim for representations that can be leveraged by classical robotics techniques. In particular, given geometric representations such as the shape~\cite{chang2015shapenet,martin2019rbo,xiang2020sapien,Xu20}, alongwith the rigid object pose~\cite{xiang2017posecnn,tremblay2018deep,sundermeyer2018implicit,wang2019normalized,chen2020learning}, articulated part pose~\cite{hu2017learning,yi2018deep,wang2019shape2motion,yan2019rpm,li2020category,xiang2020sapien,jain2020screwnet} pose, or shape functional semantics~\cite{kim2014shape2pose,hu2018functionality,hu2020predictive}, one can leverage off-the-shelf planners~\cite{miller2003automatic} or prediction systems~\cite{dexnet} developed in the robotics community to obtain action trajectories. Additionally, the ability to infer physical properties \eg material~\cite{savva2015semantically,lin2018learning}, mass~\cite{savva2015semantically,shao2017cross} \etc can further make this process accurate. However, this two-stage procedure for acting, involving a perception system that predicts the object `state', is not robust to prediction errors and makes the perception system produce richer output than possibly needed \eg we don't need the full object state to pull out a drawer. Moreover, while this approach allows an agent to precisely execute an action, it sidesteps the issue of what action needs to/can be performed in the first place \eg how does the agent understand a button can be pushed?




\mypara{Learning Affordances from Passive Observations.}
One interesting approach to allow agents to learn what actions can be taken in a given context is to leverage (passive) observations -- one can watch videos of other agents interacting with an object/scene and learn what is possible to do. This technique has been successfully used to learn scene affordances (sitting/standing)~\cite{Fouhey12}, possible contact locations~\cite{brahmbhatt2019contactdb}, interaction hotspots~\cite{nagarajan2019grounded}, or even grasp patterns~\cite{hamer2010object}. However, learning from passive observations is challenging due to several reasons \eg the learning agent may differ in anatomy thereby requiring appropriate retargeting of demonstrations. An even more fundamental concern is the distribution shift common in imitation learning -- while the agent may see examples of what can be done, it may not have seen sufficient negative examples or even sufficiently varied positive ones.

\mypara{Learning Perception by Interaction.}
Most closely related to our approach is the line of work where an agent learns to predict affordances by generating its own training data -- by interacting with the world and trying out possible actions. One important task where this approach has led to impressive results is that of planar grasping~\cite{pinto2016supersizing,khansari2020action}, where the agent can learn which grasp actions would be successful. While subsequent approaches have attempted to apply these ideas to other tasks like object segmentation~\cite{pathakCVPRW18segByInt,lohmann2020learning}, planar pushing~\cite{pinto2017learning,zeng2018learning}, or non-planar grasps~\cite{murali20206}, these systems are limited in the complexity of the actions they model. In parallel, while some methods have striven for learning more complex affordances, they do so without modeling for the low-level actions required and instead frame the task as classification with oracle manipulators~\cite{nagarajan2020exploration}. In our work, driven by availability of scalable simulation with diverse objects, we tackle the task of predicting affordances for richer interactions while also learning the low-level actions that induce the desired change.



\section{Problem Statement}
\label{sec:problem}

We formulate a new challenging problem \textbf{Where2Act} -- inferring per-pixel `actionable information' for manipulating 3D articulated objects.
As illustrated in Fig.~\ref{fig:teaser}, given a 3D shape $S$ with articulated parts (\eg the drawer and door on the cabinet), we perform per-pixel predictions for (a) \textit{where} to interact, (b) \textit{how} to interact, and (c) the interaction outcomes, under different action primitives.

In our framework, the input shape can be represented as a 2D RGB image or a 3D partial point cloud scan.
We parametrize six types of short-term primitive actions (\eg pushing, pulling) by the robot grippper pose in the $SE(3)$ space and consider an interaction successful if it interacts with the intended contact point on object validly and causes part motion to a considerable amount.

With respect to every action primitive, we predict for each pixel/point $p$ over the visible articulated parts of a 3D shape $S$ the following:
(a) an actionability score $a_p$ measuring how likely the pixel $p$ is actionable;
(b) a set of interaction proposals $\left\{R_{z|p}\in SO(3)\right\}_z$ to interact with the point $p$, where $z$ is randomly drawn from a uniform Gaussian distribution;
(c) one success likelihood score $s_{R|p}$ for each action proposal $R$.
%

\section{Method}
\label{sec:method}

We propose a learning-from-interaction approach to tackle this task. Taking as input a single RGB image or a partial 3D point cloud, we employ an encoder-decoder backbone to extract per-pixel features and design three decoding branches to predict the 'actionable information'.

\begin{figure}[t!]
    \centering
    \includegraphics[width=\linewidth]{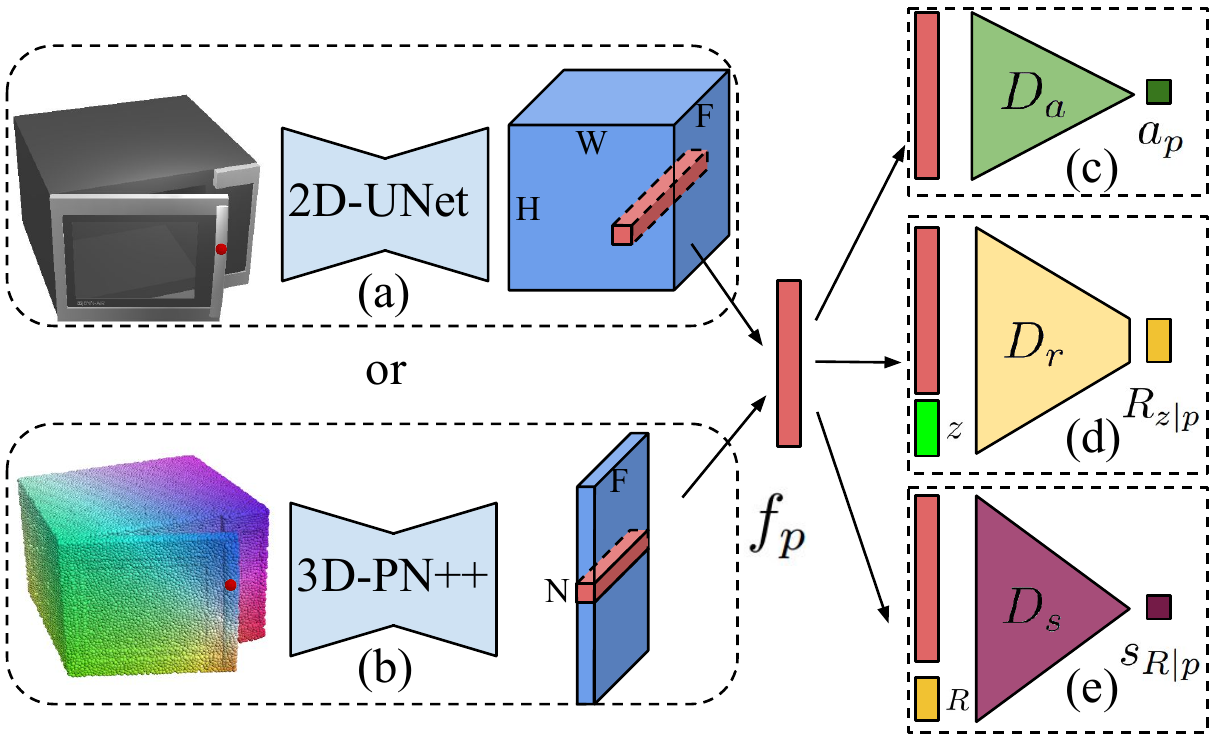}
    \caption{\titlecap{Network Architecture.}{Our network takes an 2D image or a 3D partial scan as input and extract per-pixel feature $f_p$ using (a) Unet~\cite{ronneberger2015u} for 2D images and (b) PointNet++~\cite{qi2017pointnet++} for 3D point clouds. To decode the per-pixel actionable information, we propose three decoding heads: (c) an actionability scoring module $D_a$ that predicts a score $a_p\in[0,1]$; (d) an action proposal module $D_r$ that proposes multiple gripper orientations $R_{z|p}\in SO(3)$ sampled from a uniform Gaussian random noise $z$; (e) an action scoring module $D_s$ that rates the confidence $s_{R|p}\in[0,1]$ for each proposal.}}
    \label{fig:network}
    \vspace{-4mm}
\end{figure}

\subsection{Network Modules}
Fig.~\ref{fig:network} presents an overview of the proposed method.
Our pipeline has four main components: a backbone feature extractor, an actionability scoring module, an action proposal module, and an action scoring module.
We train an individual network for each primitive action.

\vspace{1mm}
\mypara{Backbone Feature Extractor.}
We extract dense per-pixel features $\left\{f_p\right\}_p$ 
over the articulated parts. In the real-world robotic manipulation both RGB cameras or RGB-D scanners are used. Therefore, we evaluate both settings. 
For the 2D case, we use the UNet architecture~\cite{ronneberger2015u} and  implementation~\cite{Yakubovskiy:2019} with a ResNet-18~\cite{he2016deep} encoder, pretrained on ImageNet~\cite{deng2009imagenet}, and a symmetric decoder, trained from scratch, equipped with dense skip links between the encoder and decoder.
For the 3D experiments, we use PointNet++ segmentation network~\cite{qi2017pointnet++} and implementation~\cite{pytorchpointnet++} with 4 set abstraction layers with single-scale grouping for the encoder and 4 feature propagation layers for the decoder.
In both cases, we finally produce per-pixel feature $f_p\in\mathbb{R}^{128}$.

\vspace{1mm}
\mypara{Actionability Scoring Module.}
For each pixel $p$, we predict an actionability score $a_p\in[0, 1]$ indicating how likely the pixel is actionable. We employ a Multilayer Perceptron (MLP) $D_a$ with one hidden layer of size 128 to implement this module.
The network outputs one scalar $a_p$ after applying the Sigmoid function, where a higher score indicates a higher chance for successful interaction. Namely,
\begin{equation}
\vspace{-1mm}
a_p=D_a\left(f_p\right)
\vspace{-1mm}
\end{equation}

\vspace{1mm}
\mypara{Action Proposal Module.}
For each pixel $p$, we employ an action proposal module that is essentially formulated as a conditional generative model to propose high-recall interaction parameters $\left\{R_{z|p}\right\}_z$. We employ another MLP $D_r$ with one hidden layer of size 128 to implement this module.
Taking as input the current pixel feature $f_p$ and a randomly sampled Gaussian noise vector $z\in\mathbb{R}^{10}$, the network $D_p$ predicts a gripper end-effector 3-DoF orientation $R_{z|p}$ in the $SO(3)$ space
\begin{equation}
R_{z|p}=D_r\left(f_p, z\right).
\end{equation}
We represent the 3-DoF gripper orientation by the first two orthonormal axes in the $3\times 3$ rotation matrix, following the proposed 6D-rotation representation in~\cite{zhou2019continuity}.

\vspace{1mm}
\mypara{Action Scoring Module.}
For an action proposal $R$ at pixel $p$, we finally estimate a likelihood $s_{R|p}\in[0, 1]$ for the success of the interaction parametrized by tuple $(p, R)\in SE(3)$.
One can use the predicted action scores to filter out low-rated proposals, or sort all the candidates according to the predicted scores, analogous to predicting confident scores for bounding box proposals in the object detection literature.

This network module $D_s$ is also parametrized by an MLP with one hidden layer of size 128.
Given an input tuple $(f_p, R)$, we produce a scalar $s_{R|p}\in[0, 1]$,
\begin{equation}
\vspace{-1mm}
s_{R|p}=D_s\left(f_p, R\right),
\vspace{-1mm}
\end{equation}
where $s_{R|p}>0.5$ indicates a positive action proposal $R$ during the testing time.

\begin{figure}[t!]
    \centering
    \includegraphics[width=\linewidth]{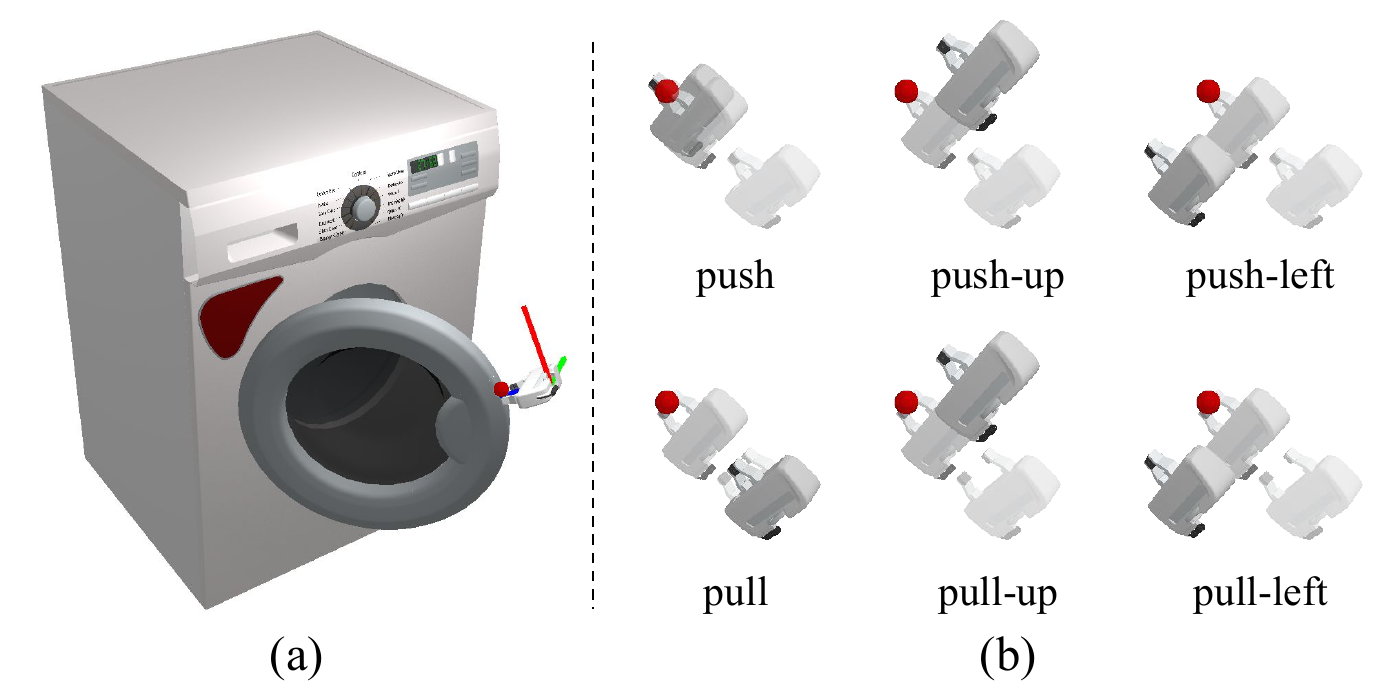}
    \caption{(a) Our interactive simulation environment: we show the local gripper frame by the red, green and blue axes, which corresponds to the leftward, upward and forward directions respectively; (b) Six types of action primitives parametrized in the $SE(3)$ space: we visualize each pre-programmed motion trajectory by showing the three key frames, where the time steps go from the transparent grippers to the solid ones, with $3\times$ exaggerated motion ranges.}
    \label{fig:env}
    \vspace{-4mm}
\end{figure}

\subsection{Collecting Training Data}
It is extremely difficult to collect human annotations for the predictions that we are pursuing.
Instead, we propose to let the agent learn by interacting with objects in simulation.
As illustrated in Fig.~\ref{fig:env} (a), we create an interactive environment using SAPIEN~\cite{xiang2020sapien} where a random 3D articulated object is selected and placed at the center of the scene. 
A flying robot gripper can then  interact with the object by specifying a position $p\in\mathbb{R}^3$ over the shape geometry surface with an end-effector orientation $R\in SO(3)$.
We consider six types of action primitives (Fig.~\ref{fig:env} (b)) with pre-programmed interaction trajectories, each of which is parameterized by the gripper pose $(p, R)\in SE(3)$ at the beginning.

We employ a hybrid data sampling strategy where we first sample large amount of offline random interaction trajectories to bootstrap the learning and then adaptively sample online interaction data points based on the network predictions for more efficient learning.

\vspace{1mm}
\mypara{Offline Random Data Sampling.}
We sample most of the training data in an offline fashion as we can efficiently sample several data points by parallelizing simulation environments across multiple CPUs.
For each data point, we first randomly sample a position $p$ over the ground-truth articulated parts to interact with.
Then, we randomly sample an interaction orientation $R\in SO(3)$ from the hemisphere above the tangent plane around $p$, oriented consistently to the positive normal direction, and try to query the outcome of the interaction parametrized by $(p, R)$.
We mark orientation $R$s from the other hemisphere as negative without trials since the gripper cannot be put inside the object volume.

In our experiments, for each primitive action type, we sample enough offline data points that give roughly $10,000$ positive trajectories to bootstrap the training.
Though parallelization allows large scale offline data collection, such random data sampling strategy is highly inefficient in querying the interesting interaction regions to obtain positive data points.
Statistics show that only 1\% data samples are positive for the \textit{pulling} primitive.
This renders a big data imbalance challenge in training the network and also hints that the most likely pullable regions occupy really small regions, which is practically very reasonable since we most likely pull out doors/drawers by their handles.

\vspace{1mm}
\mypara{Online Adaptive Data Sampling.}
To address the sampling-inefficiency of offline random data sampling, we propose to conduct online adaptive data sampling that samples more over the subregions that the network that we are learning predicts to be highly possible to be successful.

In our implementation, during training the network for the action scoring module $D_s$ with data sample $(p, R)$, we infer the action score predictions $\left\{s_{R|p_i}\right\}_i$ over all pixels $\left\{p_i\right\}_i$ on articulated parts.
Then, we sample one position $p_*$ to conduct an additional interaction trial $(p_*, R)$ according to the SoftMax normalized probability distribution over all possible interaction positions.
By performing such online adaptive data sampling, we witness an increasingly growing positive data sample rate since the network is actively choosing to sample more around the likely successful subregions.
Also, we observe that sampling more data around the interesting regions helps network learn better features at distinguishing the geometric subtleties around the small but crucial interactive parts, such as handles, buttons and knobs.

While this online data sampling is beneficial, it may lead to insufficient exploration of novel regions. Thus, in our final online data sampling procedure, we sample 50\% of data trajectories from the random data sampling and sample the other 50\% from prediction-biased adaptive data sampling.

\subsection{Training and Losses}
We empirically find it beneficial to first train the action scoring module $D_s$ and then train the three decoders jointly.
We maintain separate data queues for feeding same amount of positive and negative interaction data in each training batch to address the data imbalance issue.
We also balance sampling shapes from different object categories equally. 

\vspace{1mm}
\mypara{Action Scoring Loss.}
Given a batch of $B$ interaction data points $\left\{(S_i, p_i, R_i, r_i)\right\}_i$ where $r_i=1$ (positive) and $r_i=0$ (negative) denote the ground-truth interaction outcome, we train the action scoring module $D_s$ with the standard binary cross entropy loss
\begin{equation}
\begin{aligned}
    \mathcal{L}_s = -\frac{1}{B}\sum_i & r_i\log\left(D_s(f_{p_i|S_i}, R_i)\right)+ \\
    &(1-r_i)\log\left(1-D_s(f_{p_i|S_i}, R_i)\right).
\end{aligned}
\end{equation}

\vspace{1mm}
\mypara{Action Proposal Loss.}
We leverage the Min-of-N strategy~\cite{fan2017point} to train the action proposal module $D_r$, which is essentially a conditional generative model that maps a pixel $p$ to a distribution of possible interaction proposals $R_{z|p}$'s.
For each positive interaction data, we train $D_r$ to be able to propose one candidate that matches the ground-truth interaction orientation.
Concretely, for a batch of $B$ interaction data points $\left\{(S_i, p_i, R_i, r_i)\right\}_i$ where $r_i=1$, the Min-of-N loss is defined as 
\begin{equation}
    \mathcal{L}_r = \frac{1}{B}\sum_i \min_{j=1,\cdots,100} dist\left(\left(D_r\left(f_{p_i|S_i}; z_j\right)\right), R_i\right),
\label{eq:Lr}
\end{equation}
where $z_j$ is \textit{i.i.d} randomly sampled Gaussian vectors and $dist$ denotes a distance function between two 6D-rotation representations, as defined in~\cite{zhou2019continuity}.

\vspace{1mm}
\mypara{Actionability Scoring Loss.}
We define the `actionability' score corresponding to a pixel as the expected success rate when executing a random proposal generated by our proposal generation module $D_r$. While one could estimate this by actually executing these proposals, we note that our learned action scoring module $D_s$ allows us to directly evaluate this. We train our `actionability' scoring module to learn this expected score across proposals from $D_r$, namely, 
\begin{equation}
\begin{aligned}
    \hat{a}_{p_i|S_i}&=\frac{1}{100}\sum_{j=1,\cdots,100} D_s\left(f_{p_i|S_i}, D_r\left(f_{p_i|S_i}, z_j\right)\right); \\
    \mathcal{L}_a&=\frac{1}{B}\sum_i \left(D_a(f_{p_i|S_i}) - \hat{a}_{p_i|S_i}\right)^2.
\end{aligned}
\end{equation}
This strategy is computationally efficient since we are re-using the 100 proposals computed in Eq.~\ref{eq:Lr}. 
Also, since the action proposal network $D_r$ is optimized to cover all successful interaction orientations, the estimation $\hat{a}_{p_i|S_i}$ is expected to be approaching 1 when most of the proposals are successful and 0 when the position $p$ is not actionable (\ie all proposals are rated with low success likelihood scores).

\vspace{1mm}
\mypara{Final Loss.}
After adjusting the relative loss scales to the same level, we obtain the final objective function
\begin{equation}
    \mathcal{L}=\mathcal{L}_s+\mathcal{L}_r+100\times\mathcal{L}_a.
\end{equation}

\section{Experiments}
\label{sec:exp}
We set up an interactive simulation environment in SAPIEN~\cite{xiang2020sapien} and benchmark performance of the proposed method both qualititively and quantitatively.
Results also show that the networks learn representations that can generalize to novel unseen object categories and real-world data.

\subsection{Framework and Settings}
\label{subset:dataset_settings}
We describe our simulation environment, simulation assets and action primitive settings in details below.

\mypara{Environment.} 
Equipped with a large-scale PartNet-Mobility dataset, SAPIEN~\cite{xiang2020sapien} provides a physics-rich simulation environment that supports robot actuators interacting with 2,346 3D CAD models from 46 object categories.
Every articulated 3D object is annotated with articulated parts of interests (\eg doors, handles, buttons) and their part motion information (\ie motion types, motion axes and motion ranges).
SAPIEN integrates one of the state-of-the-art physical simulation engines NVIDIA PhysX~\cite{physx} to simulate physics-rich interaction details.

We adapt SAPIEN to set up our interactive environment for our task.
For each interaction simulation, we first randomly select one articulated 3D object, which is zero-centered and normalized within a unit-sphere, and place it in the scene.
We initialize the starting pose for each articulated part, with a 50\% chance at its rest state (\eg a fully closed drawer) and 50\% chance with a random pose (\eg a half-opened drawer).
Then, we use a Franka Panda Flying gripper with 2 fingers as the robot actuator, which has 8 degree-of-freedom (DoF) in total, including the 3 DoF position, 3 DoF orientation and 2 DoF for the 2 fingers.
The flying gripper can be initialized at any position and orientation with a closed or open gripper.
We observe the object in the scene from an RGB-D camera with known intrinsics that is mounted 5-unit far from the object, facing the object center, located at the upper hemisphere of the object with a random azimuth $[0^{\circ}, 360^{\circ})$ and a random altitude $[30^{\circ}, 60^{\circ}]$.
Fig.~\ref{fig:env} (a) visualizes one example of our simulation environment.

\mypara{Simulation Assets.} We conduct our experiments using 15 selected object categories in the PartNet-Mobility dataset, after removing the objects that are either too small (\eg pens, USB drives), requiring multi-gripper collaboration (\eg pliers, scissors), or not making sense for robot to manipulate (\eg keyboards, fans, clocks).
We use 10 categories for training and reserve the rest 5 categories only for testing, in order to analyze if the learned representations can generalize to novel unseen categories.
In total, there are 773 objects in the training categories and 199 objects in the testing ones.
We further divide the training split into 586 training shapes and 187 testing shapes, and only use the training shapes from the training categories to train our networks.
Table~\ref{tab:dataset_stats} summarizes the detailed statistics of the final data splits.

\begin{table}[t]
  \centering
  \setlength{\tabcolsep}{1pt}
  \renewcommand\arraystretch{0.5}
  \small
    \begin{tabular}{c|ccccccc}
    \toprule
        Train-Cats  & \small{All}   & \small{Box} & \small{Door}   & \small{Faucet} & \small{Kettle} & \small{Microwave}  \\ \hline
    Train-Data   & 586 & 20    & 23    & 65    & 22   & 9   \\
    Test-Data   &  187 & 8   & 12    & 19    & 7 & 3   \\
    \midrule
          &  & \small{Fridge} & \small{Cabinet} & \small{Switch} & \small{TrashCan} & \small{Window}  \\
    \hline
       &     & 32    & 270    & 53    & 52    & 40    \\
       &     & 11   & 75   & 17    & 17 & 18   \\
    \midrule[1pt]
        Test-Cats   & \small{All}   & \small{Bucket}   & \small{Pot} & \small{Safe} & \small{Table}  & \small{Washing}  \\
    \hline
    Test-Data   &  199   & 36    & 23    & 29    & 95    &  16   \\
    \bottomrule
    \end{tabular}%
    \vspace{1mm}
  \caption{We summarize the shape counts in our dataset. Here, \textit{pot} and \textit{washing} are short for kitchen pot and washing machine.}
  \label{tab:dataset_stats}
  \vspace{-4mm}
\end{table}%


\mypara{Action Settings.} We consider six types of primitive actions: \textit{pushing}, \textit{pushing-up}, \textit{pushing-left}, \textit{pulling}, \textit{pulling-up}, \textit{pulling-left}.
All action primitives are pre-programmed with hard-coded motion trajectories and parameterized by the gripper starting pose $R\in SE(3)$ in the camera space.
At the beginning of each interaction simulation, we initialize the robot gripper slightly above a surface position $p$ of interest approaching from orientation $R$.

We visualize the action primitives in Fig.~\ref{fig:env} (b).
For \textit{pushing}, a closed gripper first touches the surface and then pushes $0.05$ unit-length forward.
For \textit{pushing-up} and \textit{pushing-left}, the closed gripper moves forward by $0.04$ unit-length to contact the surface and scratches the surface to the up or left direction for $0.05$ unit-length.
For \textit{pulling}, an open gripper approaches the surface by moving forward for $0.04$ unit-length, performs grasping by closing the gripper, and pulls backward for $0.05$ unit-length.
For \textit{pulling-up} and \textit{pulling-left}, after the attempted grasping, the gripper moves along the up or left direction for $0.05$ unit-length.
Notice that the \textit{pulling} actions may degrade to the \textit{pushing} ones if the gripper grasps nothing but just touches/scratches the surface.
%

%
We define one interaction trial successful if the part that we are interacting with exhibits a considerable part motion along the intended direction.
The intended direction is the forward or backward direction for \textit{pushing} and \textit{pulling}, and is the up or left direction for the rest four directional action types.
We measure the contact point motion direction and validate it if the angle between the intended direction and the actual motion direction is smaller than $60^{\circ}$.
For thresholding the part motion magnitude, we measure the gap between the starting and end part 1-DoF pose and claim it successful if the gap is greater than $0.01$ unit-length or $0.5$ relative to the total motion range of the articulated part.

\subsection{Metrics and Baselines}
We propose two quantitative metrics for evaluating performance of our proposed method, compared with three baseline methods and one ablated version of our method.

\begin{figure*}[t!]
    \centering
    \includegraphics[width=\linewidth]{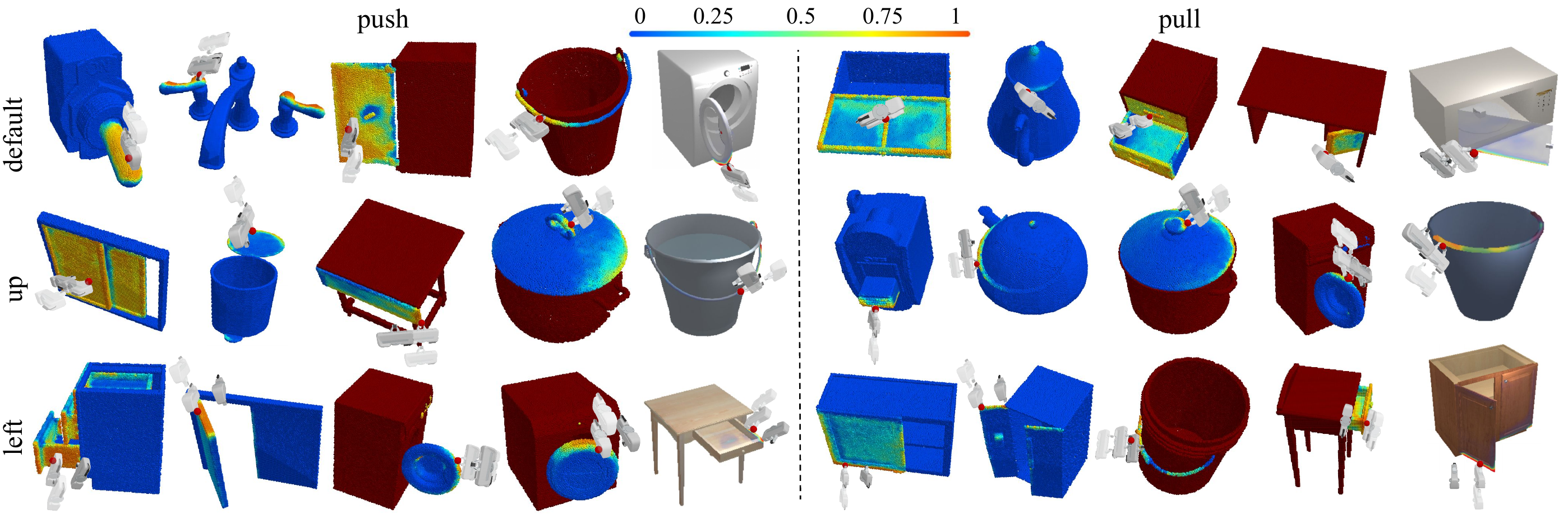}
    \caption{We visualize the per-pixel action scoring predictions over the articulated parts given certain gripper orientations for interaction. In each set of results, the left two shapes shown in blue are testing shapes from training categories, while the middle two shapes highlighted in dark red are shapes from testing categories. The rightmost columns show the results for the 2D experiments.}
    \label{fig:critic}
    \vspace{-4mm}
\end{figure*}

\mypara{Evaluation Metrics.}
A natural set of metrics is to evaluate the binary classification accuracy of the action scoring network $D_s$.
We conduct random interaction simulation trials in the SAPIEN environment over testing shapes with random camera viewpoints, interaction positions and orientations.
With random interactions, there are many more failed interaction trials than the successful ones.
Thus, we report the F-score balancing precision and recall for the positive class.

To evaluate the action proposal quality, we introduce a sample-success-rate metric $ssr$ that measures what fraction of interaction trials proposed by the networks are successful.
This metric jointly evaluates all the three network modules and mimics the final use case of proposing meaningful actions when a robot actuator wants to operate the object.
Given an input image or partial point cloud, we first use the actionability scoring module $D_a$ to sample a pixel to interact, then apply the action proposal module $D_r$ to generate several interaction proposals, and finally sample one interaction orientation according to the ratings from the action scoring module $D_s$ .
For both sampling operations, we normalize the predicted scores over all pixels or all action proposals as a probabilistic distribution and sample among the ones with absolute probability greater than $0.5$.
For the proposal generation step, we sample $100$ action proposals per pixel by randomly sampling the inputs to $D_r$ from a uniform Gaussian distribution.
For each sampled interaction proposal, we apply it in the simulator and observe the ground-truth outcome.
We define the final measure as below.
\begin{equation}
    ssr=\frac{\text{\# successful proposals}}{\text{\# total proposals}}
\end{equation}

\begin{table}[t]
  \centering
    \setlength{\tabcolsep}{5pt}
  \renewcommand\arraystretch{0.5}
  \small
\begin{tabular}{@{}llcccccc@{}}
\toprule
&&  \textsf{\footnotesize F-score (\%)} & \textsf{\footnotesize Sample-Succ (\%)}  \\
\midrule
\multirow{5}{*}{pushing}
& B-Random & 12.02 / 7.40 & 6.80 / 3.79 \\
& B-Normal & 31.94 / 17.39 & 21.72 / 11.57 \\
& B-PCPNet & 32.01 / 18.21 & 18.04 / 9.15 \\
& 2D-ours & 34.21 / 22.68 & 21.36 / 10.58 \\
& 3D-ours & \textbf{43.76} / \textbf{26.61} & \textbf{28.54} / \textbf{14.74} \\
\midrule
\multirow{5}{*}{pushing-up}
& B-Random & 4.92 / 3.31 & 2.70 / 1.62 \\
& B-Normal & 13.37 / 7.56 & 8.93 / 4.81 \\
& B-PCPNet & 15.08 / 7.50 & 8.09 / 4.86 \\
& 2D-ours & 15.35 / 8.69 & 8.70 / 5.76 \\
& 3D-ours & \textbf{21.64} / \textbf{11.20} & \textbf{12.06} / \textbf{6.56} \\
\midrule
\multirow{5}{*}{pushing-left}
& B-Random & 6.18 / 4.05 & 3.08 / 2.26 \\
& B-Normal & 18.52 / 10.72 & 11.59 / 5.72 \\
& B-PCPNet & 18.66 / 10.81 & 9.69 / 4.43 \\
& 2D-ours & 18.93 / 12.04 & 11.68 / 7.22 \\
& 3D-ours & \textbf{26.04} / \textbf{16.06} & \textbf{15.95} / \textbf{9.31} \\
\midrule
\multirow{5}{*}{pulling}
& B-Random & 2.26 / 3.19 & 1.07 / 1.55 \\
& B-Normal & 6.20 / 8.02 & 3.79 / 4.18 \\
& B-PCPNet & 7.19 / 8.57 & 4.15 / 3.71 \\
& 2D-ours & 7.04 / 8.98 & 4.07 / 4.70 \\
& 3D-ours & \textbf{10.95} / \textbf{12.88} & \textbf{7.51} / \textbf{7.85} \\
\midrule
\multirow{5}{*}{pulling-up}
& B-Random & 5.01 / 4.13 & 2.22 / 2.41 \\
& B-Normal & 13.64 / 9.40 & 8.67 / 6.08 \\
& B-PCPNet & 14.73 / 10.98 & 8.37 / 6.19 \\
& 2D-ours & 15.74 / 12.88 & 9.71 / 7.10 \\
& 3D-ours & \textbf{22.24} / \textbf{16.28} & \textbf{13.53} / \textbf{9.28} \\
\midrule
\multirow{5}{*}{pulling-left}
& B-Random & 5.83 / 4.16 & 3.06 / 2.31 \\
& B-Normal & 17.52 / 10.51 & 11.14 / 5.82 \\
& B-PCPNet & 18.89 / 11.00 & 9.12 / 5.19 \\
& 2D-ours & 16.20 / 10.16 & 10.15 / 6.05 \\
& 3D-ours & \textbf{25.22} / \textbf{14.49} & \textbf{14.25} / \textbf{7.10}   \\
\bottomrule
\vspace{1mm}
\end{tabular}
  \caption{\titlecap{Quantitative Evaluations and Comparisons.}{We compare our method to three baseline methods (\ie \textbf{B-Random}, \textbf{B-Normal} and \textbf{B-PCPNet}). In each entry, we report the numbers evaluated over the testing shapes from training categories (before slash) and the shapes from the test categories (after slash). We use \textbf{3D-} and \textbf{2D-} to indicate the data input modalities. The baseline methods are not sensitive to the input kinds. We observe that \textbf{3D-ours} achieves the best performance.}}
  \label{tab:results}
  \vspace{-4mm}
\end{table}%
\begin{table}[t]
  \centering
    \setlength{\tabcolsep}{5pt}
  \renewcommand\arraystretch{0.5}
  \small
\begin{tabular}{@{}llcccccc@{}}
\toprule
&&  \textsf{\footnotesize F-score (\%)} & \textsf{\footnotesize Sample-Succ (\%)}  \\
\midrule
\multirow{2}{*}{pushing}
& Ours w/o OS & 40.54 / 25.66 & 25.18 / 11.76 \\
& Ours & \textbf{43.76} / \textbf{26.61} & \textbf{28.54} / \textbf{14.74} \\
\midrule
\multirow{2}{*}{pushing-up}
& Ours w/o OS & 21.03 / \textbf{11.57} & \textbf{12.88} / 6.43 \\
& Ours & \textbf{21.64} / 11.20 & 12.06 / \textbf{6.56} \\
\midrule
\multirow{2}{*}{pushing-left}
& Ours w/o OS & 24.71 / 14.91 & 14.12 / 7.02 \\
& Ours & \textbf{26.04} / \textbf{16.06} & \textbf{15.95} / \textbf{9.31} \\
\midrule
\multirow{2}{*}{pulling}
& Ours w/o OS & 10.28 / 12.09 & 5.62 / 5.87 \\
& Ours & \textbf{10.95} / \textbf{12.88} & \textbf{7.51} / \textbf{7.85} \\
\midrule
\multirow{2}{*}{pulling-up}
& Ours w/o OS & 20.51 / 13.70 & 12.18 / 7.96 \\
& Ours & \textbf{22.24} / \textbf{16.28} & \textbf{13.53} / \textbf{9.28} \\
\midrule
\multirow{2}{*}{pulling-left}
& Ours w/o OS & 23.41 / \textbf{15.07} & 14.23 / 6.81 \\
& Ours & \textbf{25.22} / 14.49 & \textbf{14.25} / \textbf{7.10}   \\
\bottomrule
\end{tabular}
\vspace{1mm}
  \caption{\titlecap{Ablation Study.}{We compare our method to an ablated version, where we remove the online adaptive sampling. It is clear to see that using \textit{online data sampling} (OS) helps in most cases.}}
  \label{tab:ablaresults}
  \vspace{-4mm}
\end{table}%

\mypara{Baselines and Ablation Study.}
Since we are the first to propose and formulate the task, there is no previous work for us to compare with.
To validate the effectiveness of the proposed method and provide benchmarks for the proposed task, we compare to three baseline methods and one ablated version of our method:
\begin{itemize}
    \vspace{-2mm}
    \item \textbf{B-Random}: a random agent that always gives a random proposal or scoring;
    \vspace{-2mm}
    \item \textbf{B-Normal}: a method that replaces the feature $f_p$ in our method with the 3-dimensional ground-truth normal, with the same decoding heads, losses and training scheme as our proposed method; 
    \vspace{-2mm}
    \item \textbf{B-PCPNet}: a method that replaces the feature $f_p$ in our method with predicted normals and curvatures, which are estimated using  PCPNet~\cite{guerrero2018pcpnet} on 3D partial point cloud inputs, with the same decoding heads, losses and training scheme as our proposed method;
    \vspace{-2mm}
    \item \textbf{Ours w/o OS}: an ablated version of our method that removes the online adaptive data sampling strategy and only samples online data with random exploration. We make sure that the total number of interaction queries is the same as our method for a fair comparison.
\end{itemize}

Among baseline methods, \textbf{B-Random} presents lower bound references for the proposed metrics, while \textbf{B-Normal} is designed to validate that our network learns localized but interaction-oriented features, rather than simple geometric features such as normal directions.
\textbf{B-PCPNet} further validates that our network learns geometric features more than local normals and curvatures.
An ablated version \textbf{Ours w/o OS} further proves the improvement provided by the proposed \textit{online adaptive data sampling} (OS) strategy.

\subsection{Results and Analysis}
Table~\ref{tab:results} presents quantitative comparisons of our method to the three baselines, where we observe that \textbf{3D-Ours} performs the best.
Our network learns localized but interaction-oriented geometric features, performing better than \textbf{B-Normal} and \textbf{B-PCPNet} which only use normals and curvatures as features.
Though lacking of explicit 3D information and thus performing worse than the 3D networks, we observe competitive results from the 2D-version \textbf{2D-Ours}.
Our networks also learn representations that generalize successfully to unseen novel object categories.
The ablation study shown in Table~\ref{tab:ablaresults} further validates that the \textit{online data sampling} (OS) strategy helps boost the performance.

\begin{figure}[t!]
    \centering
    \includegraphics[width=\linewidth]{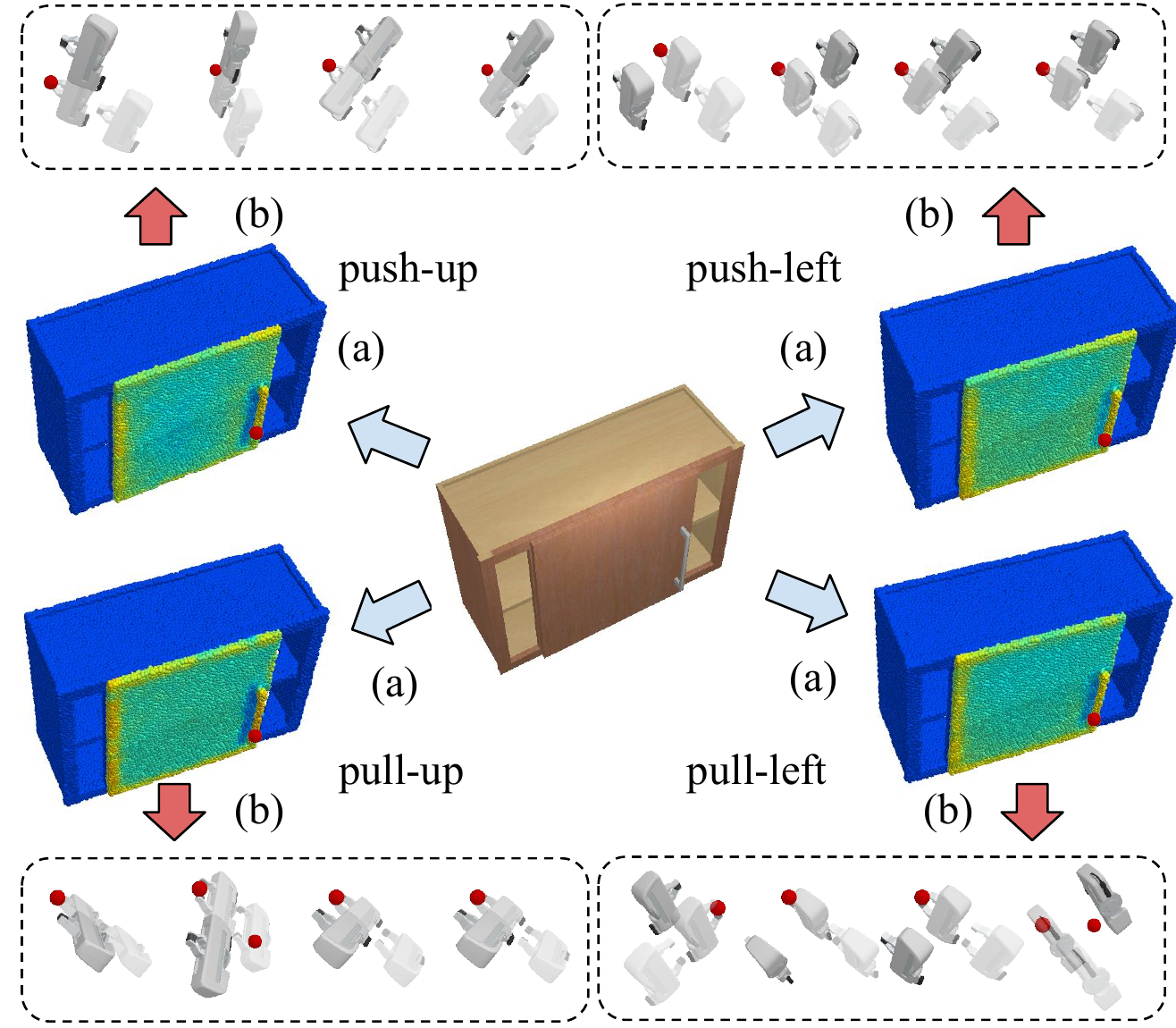}
    \caption{We visualize (a) the actionability scoring and (b) the action proposal predictions on an example cabinet with a door that can be slipped to open. We show the top-4 rated proposals.}
    \label{fig:actionability_proposal}
    \vspace{-4mm}
\end{figure}


We visualize the predicted action scores in Fig.~\ref{fig:critic}, where we clearly see that given different primitive action types and gripper orientations, our network learns to extract geometric features that are action-specific and gripper-aware.
For example, for \textit{pulling}, we predict higher scores over high-curvature regions such as part boundaries and handles, while for \textit{pushing}, almost all flat surface pixels belonging to a pushable part are equally highlighted and the pixels around handles are reasonably predicted to be not pushable due to object-gripper collisions.
For the directional interaction types, it is obvious to see that the action direction is of important consideration to the predictions.
For instance, the \textit{pushing-left} agent learns to scratch the side surface pixels of the cabinet drawers to close them (third-row, the leftmost column) and the \textit{pulling-up} one learns to lift up the handle of a bucket by grasping it and pulling up (second-row, the rightmost column).

We illustrate the estimated actionability scores over the articulated part for the six action primitives in Fig.~\ref{fig:teaser} and Fig.~\ref{fig:actionability_proposal}. 
We obverse that the door/drawer handles and part boundaries are highlighted, especially for \textit{pulling} and \textit{pulling-up}, where reasonable interaction proposals are produced.
Fig.~\ref{fig:teaser} clearly shows the different actionability predictions over the door pixels, where the door surface pixels are in general pushable, while only the handle part is pullable.
Fig.~\ref{fig:actionability_proposal} presents comparisons among the four directional interaction types.
We observe similar actionability predictions for \textit{pushing-up} and \textit{pushing-left} but different orientation proposals for interacting with the same pixel.
Interestingly, comparing \textit{pulling-up} and \textit{pulling-left}, we see that the operation of grasping is in function for \textit{pulling-up}, making it more actionable than \textit{pulling-left} when attempting to slide open the cabinet door.
We present more results in the supplementary.

\begin{figure}[t!]
    \centering
    \includegraphics[width=\linewidth]{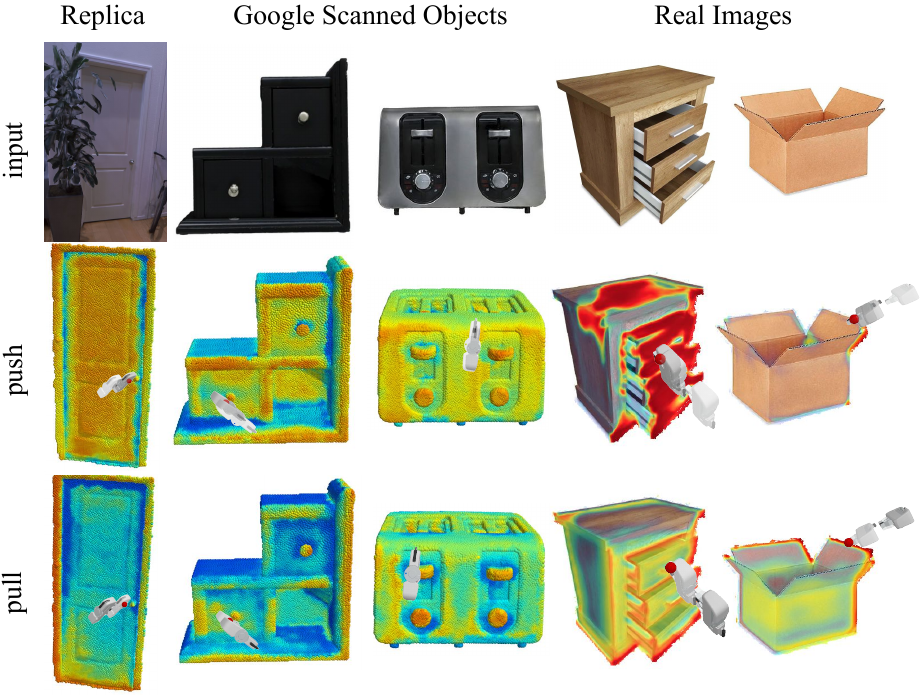}
    \caption{We visualize our action scoring predictions given certain gripper orientations over three real-world 3D scans from the Replica dataset~\cite{straub2019replica} and Google Scanned Objects~\cite{gso1,gso2}, as well as on two 2D real images~\cite{realimages}.
    Results are shown over all pixels because of no access to the articulated part masks. Though there is no guarantee for the predictions over pixels outside the articulated parts, the results make sense if we allow motion for the entire objects. }
    \label{fig:real_data}
    \vspace{-4mm}
\end{figure}

\subsection{Real-world Data}
We directly applied our networks trained on synthetic data to real-world data.
Fig.~\ref{fig:real_data} presents our predictions of the action scoring module on real 3D scans and 2D images, which shows promising results that our networks transfer the learned actionable information to real-world data.

\section{Conclusion}
\label{sec:conclusion}
We formulate a new challenging task to predict per-pixel actionable information for manipulating articulated 3D objects.
Using an interactive environment built upon SAPIEN and the PartNet-Mobility dataset, we train neural networks that map pixels to actions: for each pixel on a articulated part of an object, we predict the actionability of the pixel related to six primitive actions and propose candidate interaction parameters.
We present extensive quantitative evaluations and qualitative analysis of the proposed method.
Results show that the learned knowledges are highly localized and thus generalizable to novel unseen object categories.

\mypara{Limitations and Future Works.} We see many possibilities for future extensions.
First, our network takes single frame visual input, which naturally introduces ambiguities for the solution spaces if the articulated part mobility information cannot be fully determined from a single snapshot. 
Second, we limit our experiments to six types of action primitives with hard-coded motion trajectories. One future extension is to generalize the framework to free-form interactions.
Finally, our method does not explicitly model the part segmentation and part motion axis, which may be incorporated in the future works to further improve the performance.


\section*{Acknowledgements}
This work was supported primarily by Facebook during Kaichun's internship, while also by NSF grant IIS-1763268, a Vannevar Bush faculty fellowship, and an Amazon AWS ML award.
We thank Yuzhe Qin and Fanbo Xiang for providing helps on setting up the SAPIEN environment.

{\small
\bibliographystyle{ieee_fullname}
\bibliography{ref}
}

\appendix

\section{Framework and Settings: More Details}
For our interactive simulation environment based on SAPIEN, we use the same set of simulation parameters for all interaction trials. 
We will release our simulation environment and full toolkits for the best reproducibility of our work and supporting future research.
Besides the information provided in the main paper (Sec.~\ref{subset:dataset_settings}, \textbf{Environment}), we describe more detailed settings in our framework.

For general simulation settings, we use frame rate 500 fps, tolerance length 0.001, tolerance speed 0.005, solver iterations 20 (for constraint solvers related to joints and contacts), with Persistent Contact Manifold (PCM) disabled (for better simulation stability), with disabled sleeping mode (\ie no locking for presumably still rigid bodies in simulation), and all the other settings as default in SAPIEN release.
Following the SAPIEN suggested criterion, we also disable collision simulation between each articulated part to its direct parent node, due to usually omitted or inaccurate geometry modeling details at joint positions for ShapeNet~\cite{chang2015shapenet} models.

For physical simulation, we use the standard gravity 9.81, static friction coefficient 4.0, dynamic friction coefficient 4.0, and restitution coefficient 0.01.
For the object articulation dynamics simulation, we use stiffness 0 and damping 10.
And for the robot gripper, we use stiffness 1000 and damping 400 for the free 6-DoF robot hand motion, while we use stiffness 200 and damping 60 for the gripper fingers.

For the rendering settings, we use an OpenGL-based rasterization rendering for the fast speed of simulation.
We set three point lights around the object (one at the front, one from back-left and one from back-right) for lighting the scene, with mild ambient lighting as well.
The camera is set to have near plane 0.1, far plane 100, resolution 448, and field of view 35$^{\circ}$.
For RGB image inputs, we downsample the obtained $448\times 448$ images to $224\times 224$ before feeding to the UNet backbone.
For 3D partial point cloud scan inputs, we back-project the depth image into a foreground point cloud, by rejecting the far-away background depth pixels, and then perform furthest point sampling to get a 10K-size point cloud scan.

Each articulated object is approximated by convex hulls using the V-HACD algorithm~\cite{mamou2016volumetric} at the part level before simulation.
The object is assumed to be fixed at its root part, with only its articulated parts movable.
After loading each object to the scene and randomly initializing the starting articulated part poses, there are chances that the parts are not still due to the gravity or collision forces. Thus, we wait for 20K time steps to simulate the final rest part states until the parts are still for 5K steps, or this interaction is invalidated.
We also remove interaction trials if the object parts have initial collisions, by detecting impulses bigger than 0.0001, due to unstable simulation outcomes.

\section{Interaction Trials: More Details}
In the main paper (Sec.~\ref{subset:dataset_settings}, \textbf{Action Settings}, the second paragraph), we detailedly defined our pre-programmed motion trajectories for the six types of action primitives.
In the supplementary video, we further illustrate the interaction demonstrations in action.
Below, we describe how the robot is driven to follow the desired motion trajectories and how to collect successful interaction trials.

The dynamics of the articulated objects and robot gripper is simulated using a velocity controller, equipped with the NVIDIA PhysX internal PID controller, that drives the gripper from one position to another, while the high-level trajectory planning is done by a simple kinematic-level computed interpolation between the starting and end end-effector poses with known gripper configurations.
The robot gripper can be intialized as closed (perfectly touched) or open (0.08 unit-length apart).

For an interaction trial to be considered successful, it not only needs to cause considerable part motion along intended direction, as described in the main paper (Sec.~\ref{subset:dataset_settings}, \textbf{Action Settings}, the last paragraph), but has to be a valid interaction beforehand.
First, the interaction direction should belong to the positive hemisphere along with the surface normal direction.
Second, the robot gripper should have no collision or contact with the object at the initial state. Otherwise, we treat this interaction trial to be failed without simulation.
Finally, for the \textit{pushing} action primitives, we require that the first-time contact happens between the robot closed gripper and the target articulated part, to remove the case that the robot is pushing the other parts if multiple parts are very close to each other.
It is also invalid if the robot hand, instead of the fingers, to first touch the part.
We do not put this constraint for the \textit{pulling} primitives, as the open gripper may touch the other parts first and then grasp the target part.
For these invalid interactions, we mark them as false data points without measuring the part motion.

\section{Training Details and Computational Timing}
We use learning rate 0.001 and Adam optimizer. 
There is no image-based data augmentation. For 3D scans, we randomly down-sample point cloud inputs for augmentation.
The input shapes are $224\times224$ for images and $10000\times3$ for point clouds.
Each simulated interaction takes about 2-8s. With parallel computation, it takes 3-4 days to collect all offline interactions.
The training takes 0.8s for each iteration and 4-5 days until convergence. 
As inference is a simple forward pass, it only takes 4ms to infer 
for a batch of 32 RGB/depth images.

\section{Simulation Assets: Visualization}
\begin{figure*}[t!]
    \centering
    \includegraphics[width=0.9\linewidth]{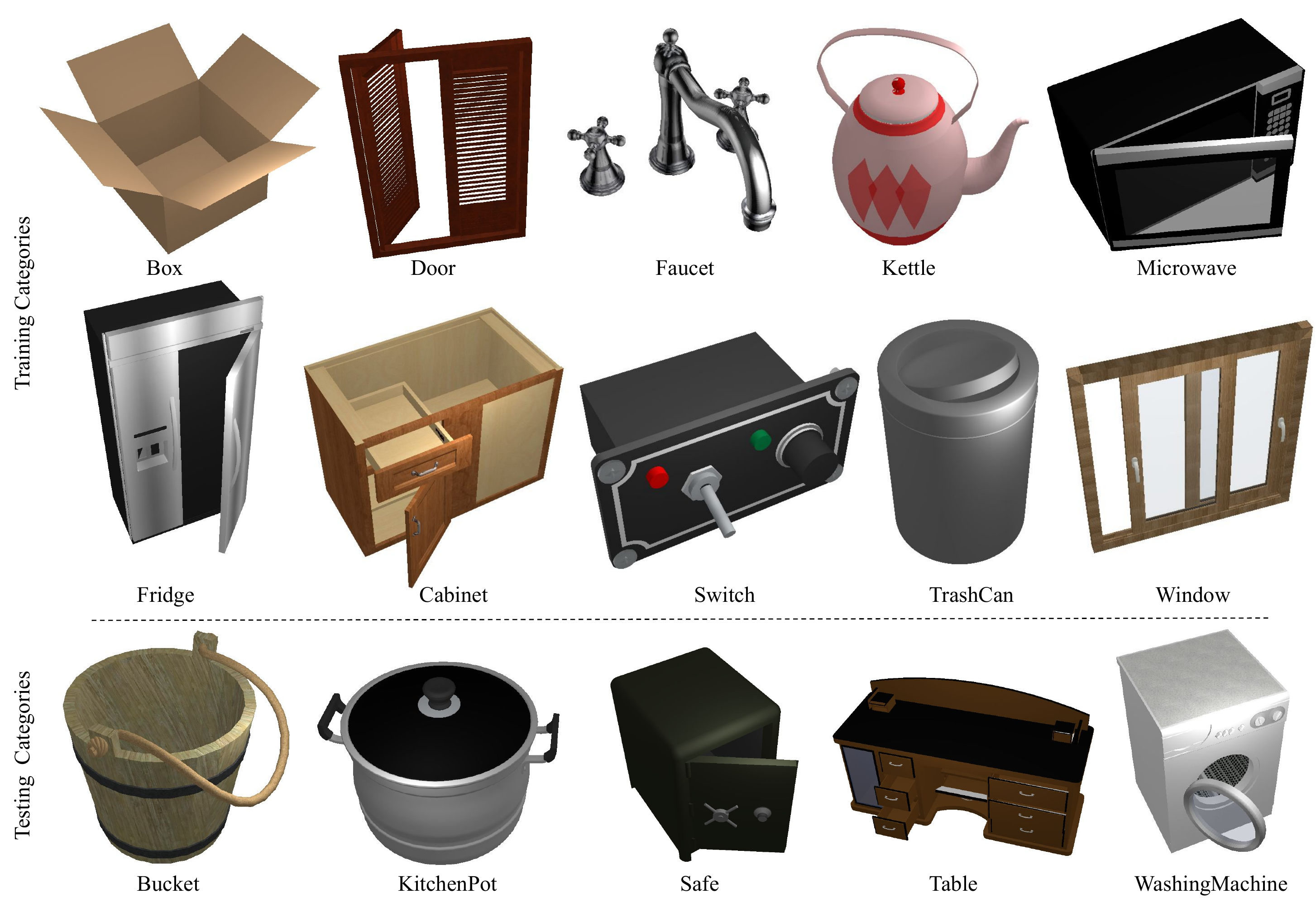}
    \caption{\titlecap{Simulation Assets Visualization.}{We visualize one example for each of the 15 object categories we use in our work.}}
    \label{supp_fig:assets}
\end{figure*}

In Fig.~\ref{supp_fig:assets}, we visualize one example for each of the 15 object categories from SAPIEN~\cite{xiang2020sapien} we use in our work.

\section{More Results on Real-world Data}
\begin{figure*}[t!]
    \centering
    \includegraphics[width=0.9\linewidth]{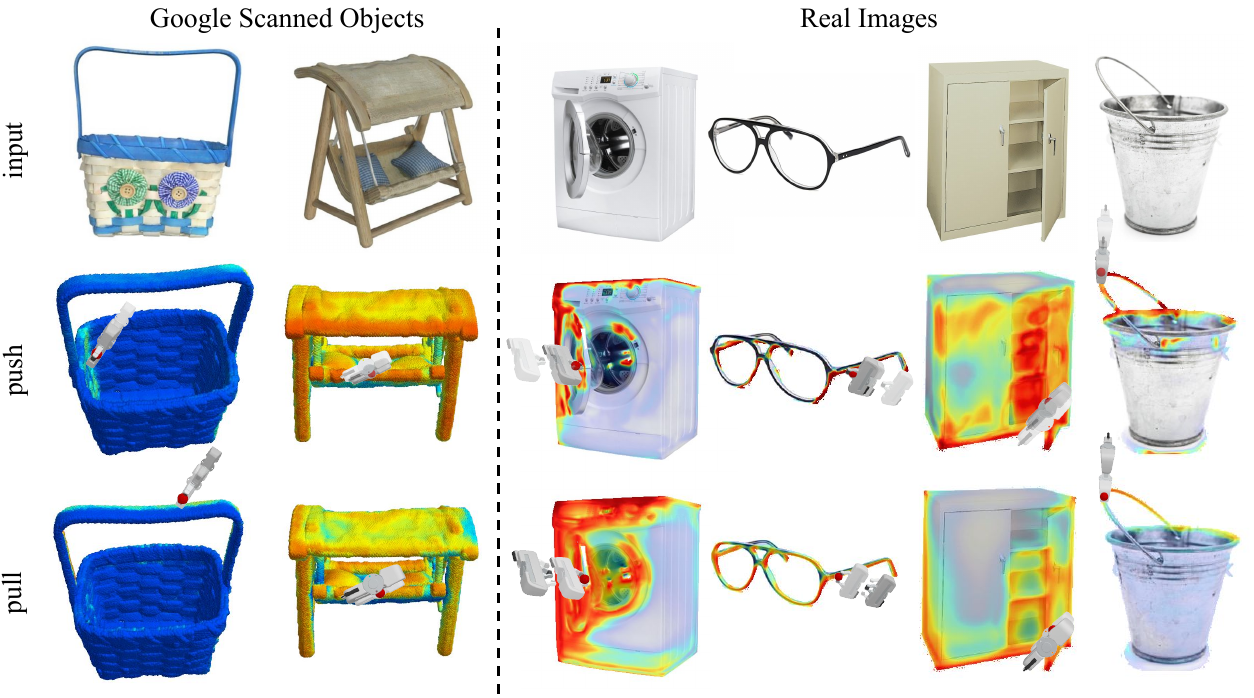}
    \caption{\titlecap{More Results on Real-world Data.}{We present more results on real-world data that augment Fig. 6 in the main paper. We use 3D real object scans from Google Scanned Objects~\cite{supp_gso2,supp_gso3} and 2D real images from the web~\cite{moreimages}. Here, results are shown over all pixels since we have no access to the articulated part masks. Though there is no guarantee for the predictions over pixels outside the articulated parts, the results make sense if we allow motion for the entire objects.}}
    \label{supp_fig:realdata}
    \vspace{-4mm}
\end{figure*}

In Fig.~\ref{supp_fig:realdata}, we visualize more results for directly applying our networks over real-world data.

\section{Actionability Scoring Predictions:\\ More Result Visualization}
\begin{figure*}[t!]
    \centering
    \includegraphics[width=\linewidth]{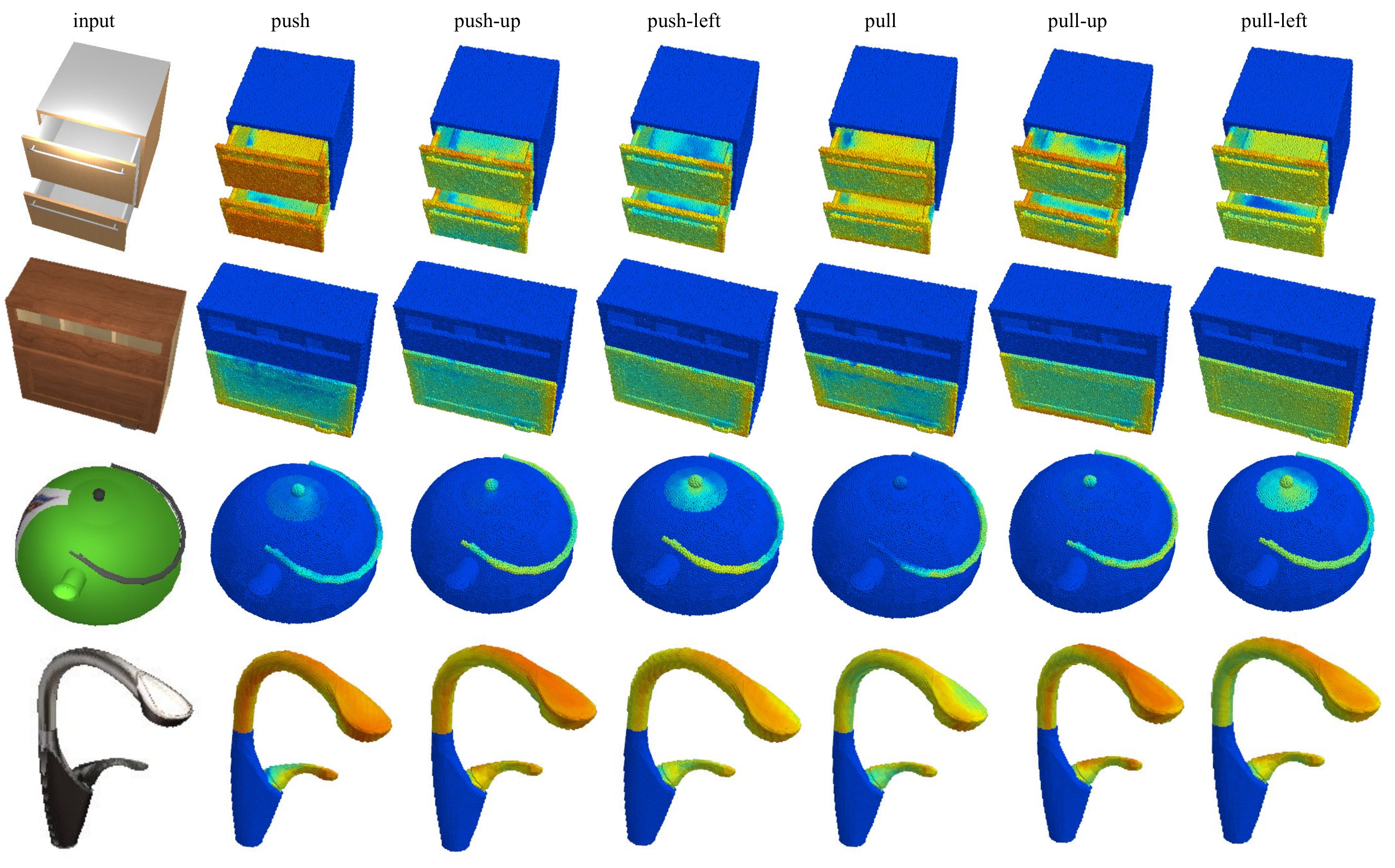}
    \includegraphics[width=\linewidth]{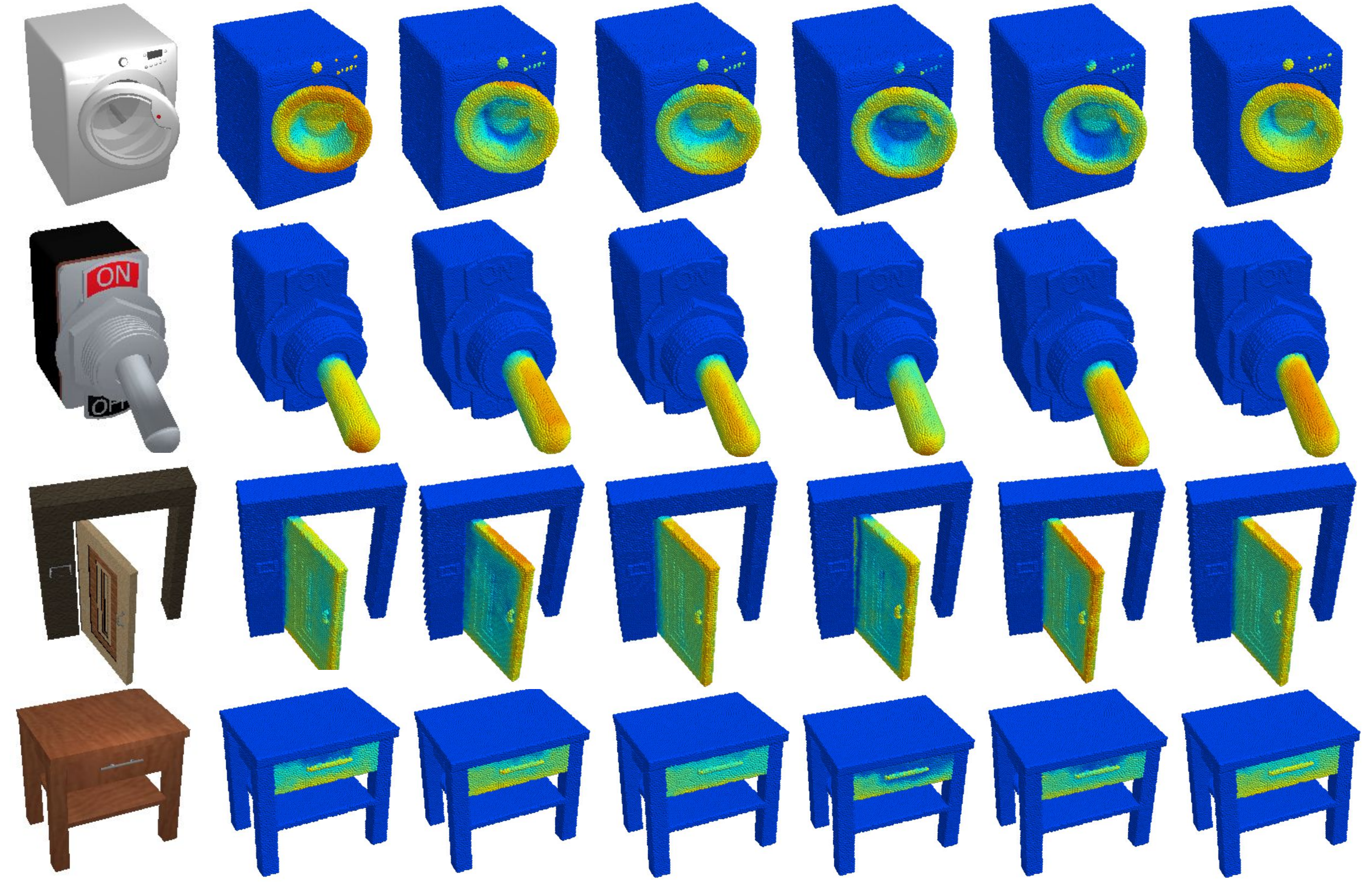}
    \caption{\titlecap{Actionability Scoring Predictions.}{We visualize more example predictions of the actionability scoring module for the six types of action primitives.}}
    \label{supp_fig:actionability_more}
    \vspace{-4mm}
\end{figure*}

In Fig.~\ref{supp_fig:actionability_more}, we visualize more example results of the actionability scoring module for the six types of action primitives.

\section{Action Proposal Predictions:\\ More Result Visualization}
\begin{figure*}[t!]
    \centering
    \includegraphics[width=\linewidth]{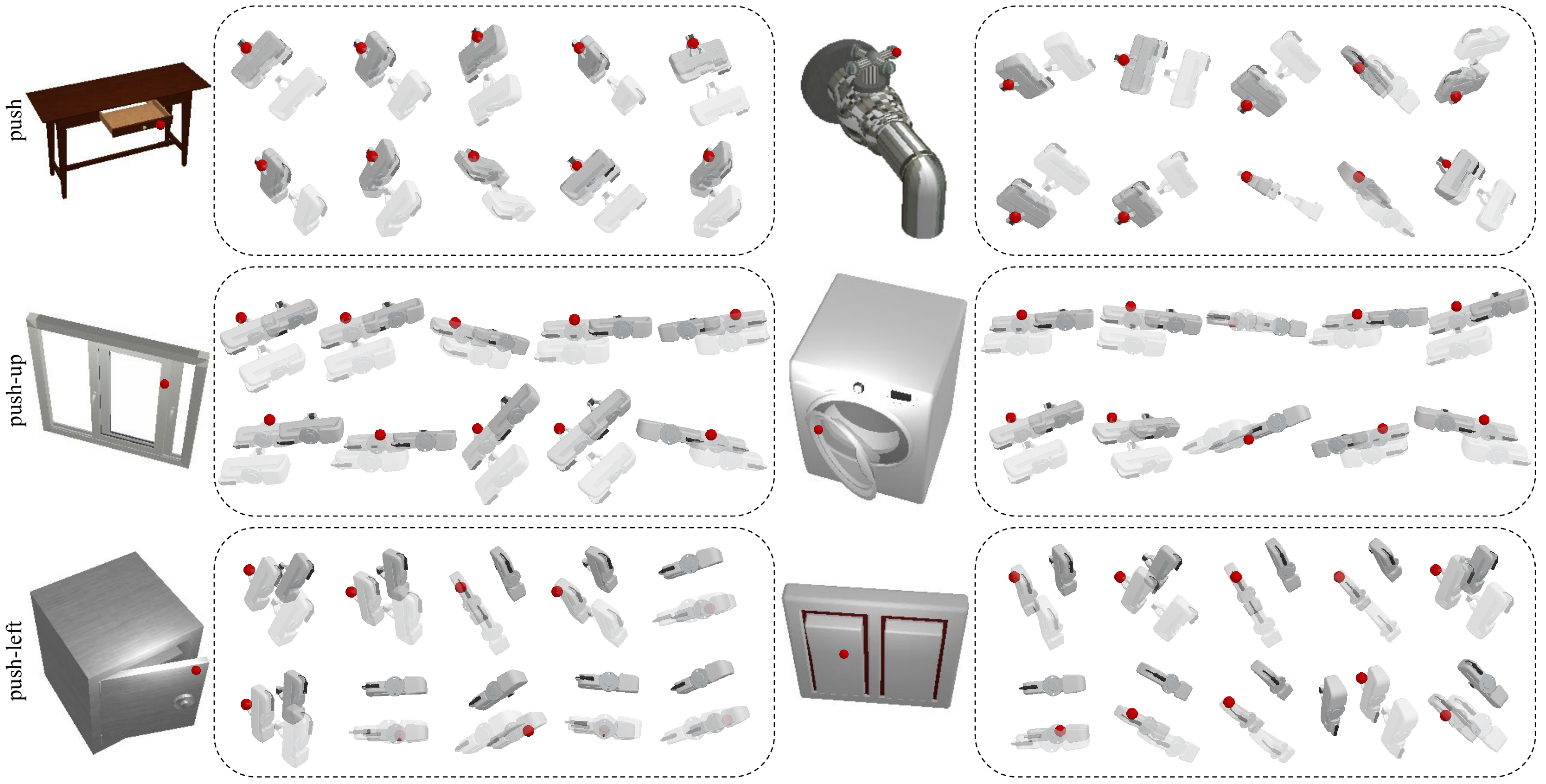}
    \includegraphics[width=\linewidth]{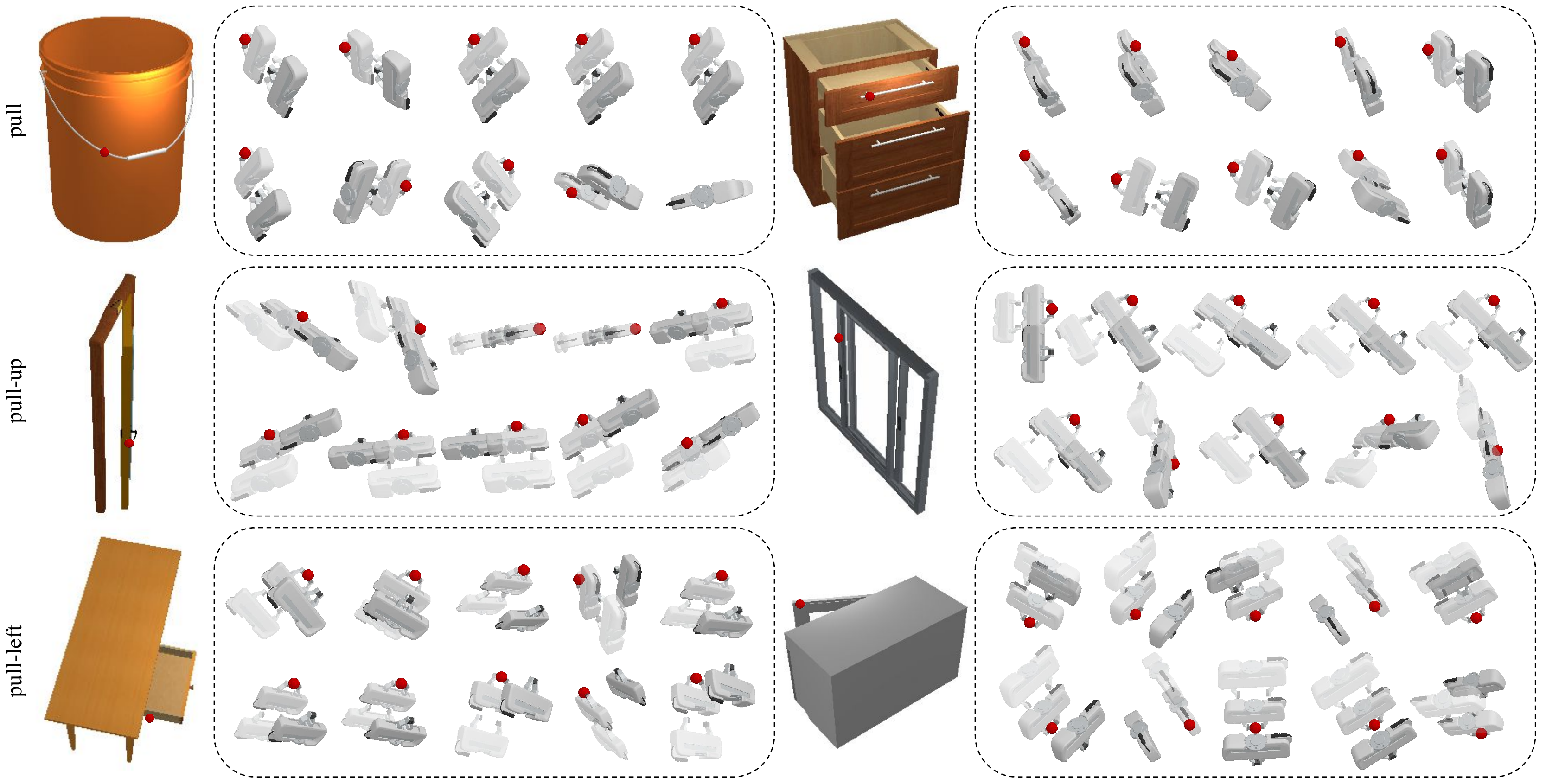}
    \includegraphics[width=\linewidth]{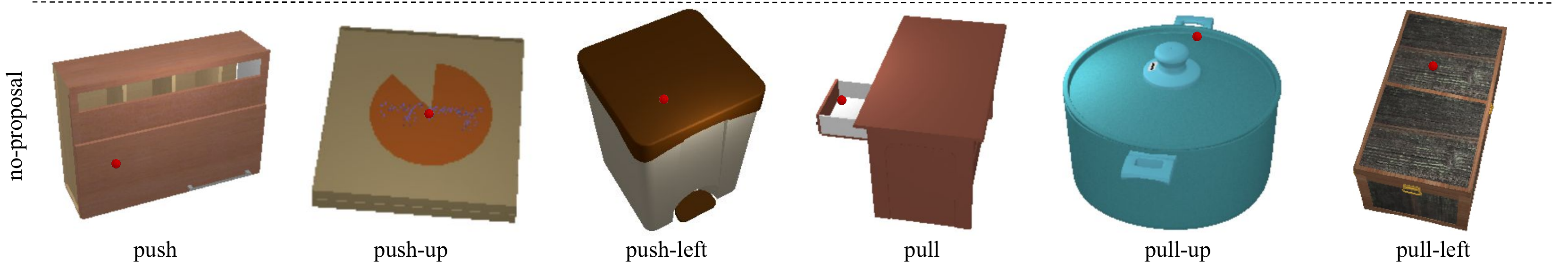}
    \caption{\titlecap{Action Proposal Predictions.}{We visualize the top-10 action proposal predictions (motion trajectories are $3\times$ exaggerated) for some example testing shapes under each action primitive. The bottom row presents the cases that no action proposal is predicted, indicating that these pixels are not actionable under the action primitives.}}
    \label{supp_fig:act_prop_more}
    \vspace{-4mm}
\end{figure*}

In Fig.~\ref{supp_fig:act_prop_more}, we visualize more action proposal predictions on example shapes for each action primitive.

\section{Failure Cases: Discussion and Visualization}
\begin{figure}[t]
    \centering
    \includegraphics[width=\linewidth]{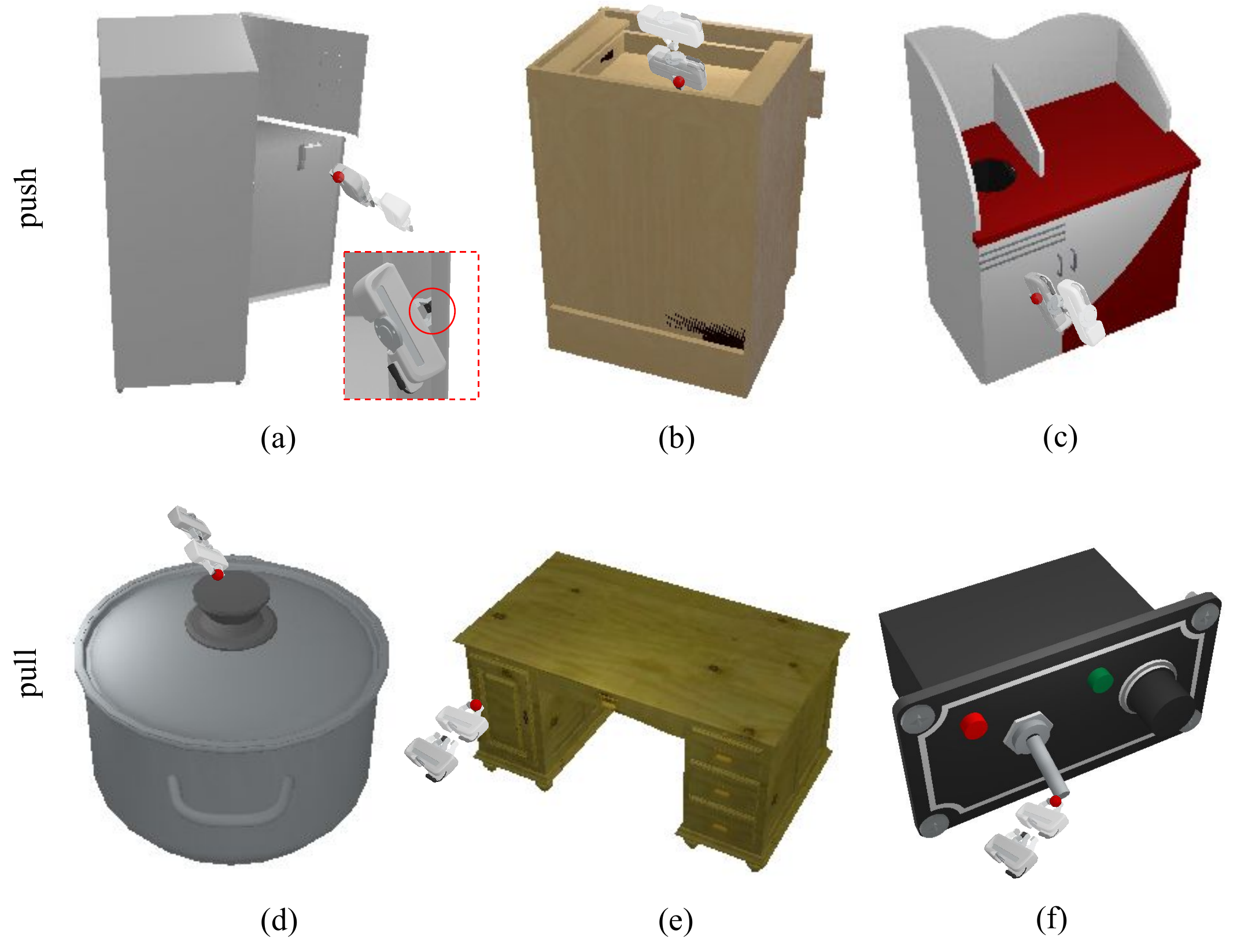}
    \caption{\titlecap{Failure Cases.}{We visualize some interesting failure cases, which demonstrate the difficulty of the task and some ambiguous cases that are hard for robot to figure out. For the \textit{pushing} action, we show (a) an example of gripper-object invalid collision at the initial state, thus leading to failed interaction, though the interaction direction seems to be successful; (b) a failed interaction due to the fact that the part motion does not surpass the required amount $0.01$ since the interaction direction is quite orthogonal to the drawer surface; and (c) a case that the door is fully closed and thus not pushable, though there are cases that the doors can be pushed inside in the dataset. For the \textit{pulling} action, we present (d) a failed grasping attempt since the gripper is too small and the pot lid is too heavy; (e) a case illustrating the intrinsic ambiguity that the robot does not know from which side the door can be opened; and (f) a failed pulling attempt as the switch toggle already reaches the allowed maximal motion range.}}
    \label{supp_fig:failure_cases}
    \vspace{-4mm}
\end{figure}

We present some interesting failure cases in Fig.~\ref{supp_fig:failure_cases}.
Please see the figure caption for detailed explanations and discussions.
From these examples, we see the difficulty of the task.
Also, given the current problem formulation, there are some intrinsically ambiguous cases that are generally hard for robot to figure out from a single static snapshot.

\end{document}